\documentclass{article}

\usepackage{microtype}
\usepackage{graphicx}
\usepackage{subfigure}
\usepackage{booktabs} 

\usepackage{hyperref}

\usepackage[accepted]{icml2024}

\usepackage{amsmath}
\usepackage{amssymb}
\usepackage{mathtools}
\usepackage{amsthm}

\usepackage[capitalize,noabbrev]{cleveref}

\theoremstyle{plain}
\newtheorem{theorem}{Theorem}[section]

\theoremstyle{definition}
\newtheorem{definition}[theorem]{Definition}

\theoremstyle{remark}

\usepackage[disable,textsize=tiny]{todonotes}

\usepackage{multirow}
\usepackage{colortbl}

\newcommand{\x}{\boldsymbol{x}}

\newcommand{\tha}{\boldsymbol{\theta}}
\newcommand{\Lm}{\mathcal{L}}
\newcommand{\Sm}{\mathcal{S}}
\newcommand{\Tm}{\mathcal{T}}

\icmltitlerunning{Learning Modality Knowledge Alignment for Cross-Modality Transfer}

\begin{document}

\twocolumn[
\icmltitle{Learning Modality Knowledge Alignment for Cross-Modality Transfer}



\icmlsetsymbol{equal}{*}

\begin{icmlauthorlist}
\icmlauthor{Wenxuan Ma}{bit}
\icmlauthor{Shuang Li}{bit}
\icmlauthor{Lincan Cai}{bit}
\icmlauthor{Jingxuan Kang}{sch}
\end{icmlauthorlist}

\icmlaffiliation{bit}{Beijing Institute of Technology, Beijing, China}
\icmlaffiliation{sch}{University of Illinois Urbana-Champaign, USA}

\icmlcorrespondingauthor{Shuang Li}{shuangli@bit.edu.cn}

\icmlkeywords{Machine Learning, ICML}

\vskip 0.3in
]



\printAffiliationsAndNotice{}  

\begin{abstract}
Cross-modality transfer aims to leverage large pretrained models to complete tasks that may not belong to the modality of pretraining data. Existing works achieve certain success in extending classical finetuning to cross-modal scenarios, yet we still lack understanding about the influence of modality gap on the transfer. In this work, a series of experiments focusing on the source representation quality during transfer are conducted, revealing the connection between larger modality gap and lesser knowledge reuse which means ineffective transfer. We then formalize the gap as the knowledge misalignment between modalities using conditional distribution $P(Y|X)$. Towards this problem, we present \textbf{Mo}dality k\textbf{N}owledge \textbf{A}lignment (MoNA), a meta-learning approach that learns target data transformation to reduce the modality knowledge discrepancy ahead of the transfer. Experiments show that out method enables better reuse of source modality knowledge in cross-modality transfer, which leads to improvements upon existing finetuning methods
\end{abstract}

\section{Introduction}
\label{sec:intro}
\begin{figure}[t]
    \centering
    \includegraphics[width=0.485\textwidth]{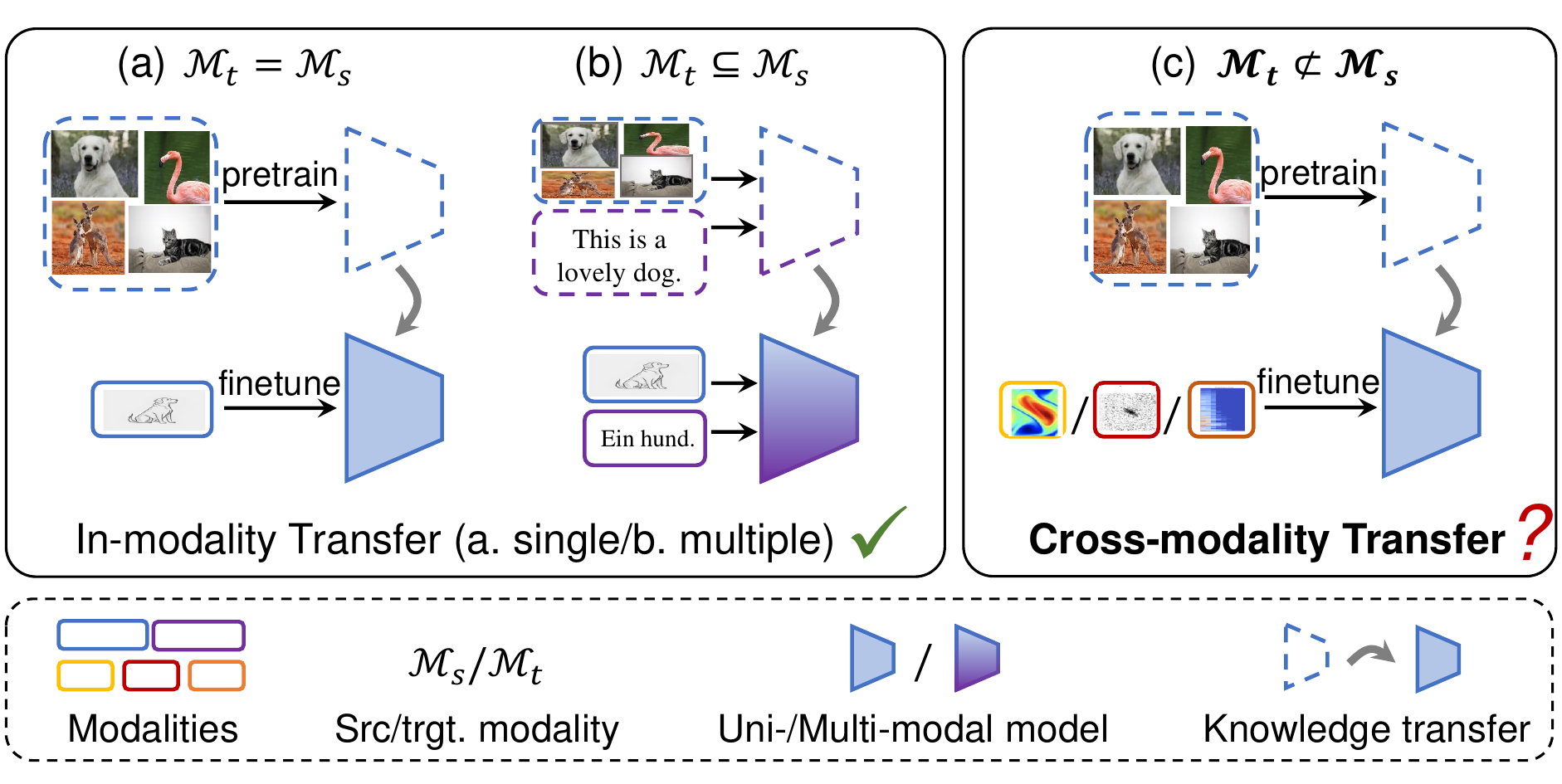}
    \vspace{-10pt}
    \caption{Comparison between in-modality finetuning (a)(b) and cross-modal finetuning (c). Both unimodal and multimodal finetuning are considered to be in-modality finetuning because the target modality is in the scope of the pretrained model's modality. In contrast, cross-modal finetuning exploits the pretrained model on target modalities that the pretrained model is not trained on.}
    \label{Fig:intro}
\end{figure}

Transferring knowledge from past experience to new tasks is a fundamental ability of human intelligence~\cite{survey,DAN-PAMI,survey_tl2020}. Such an ability to acquire and reuse knowledge is continuously pursued in machine learning community, aiming to build artificial intelligence systems that predicts more accurately and learns more data-efficiently. Today, as large fundation models that are trained on massive data are widely available~\cite{lvm,llama2,llava}\nocite{gpt3,clip,flamingo,foundation_model}, using such pretrained model as powerful feature extractor for new tasks has become a common practice of tranfer learning~\cite{peft_nature,survey_instruction_tuning}\nocite{transferable,rcnn,lora}. Naturally, the pretrained model and the downstream task come from the same modality, e.g., the model is a vision transformer pretrained on ImageNet~\cite{vit} and the task is CIFAR-100 classification~\cite{cifar100}. However, recent stuides have been attempt to broaden this boundary to cross-modality transfer, using vision transformer for audio classification~\cite{lavish}, and finetuning language model for tabular data~\cite{lift,lm4tabular}. Fig.~\ref{Fig:intro} illustrates the difference between in-modality and cross-modality transfer.

The motivation of such cross-modal transfer is easy to comprehend, especially when the target modality is data scarce. Scientific tasks like electrocardiogram classification~\cite{ecg} and protein distance prediction~\cite{psicov} find difficulties in collecting large amount of training data, and further requires much expensive annotating costs from human experts. In such cases, it is desirable to leverage the pretrained model from other modalities like vision and language, in which data are easier to collect, to help the target modality tasks. However, the cross-modal transfer is not as straightforward as the in-modality transfer due to two challenges: 1) the input space and the label space are different across modalities, and 2) the knowledge required for addressing tasks in different modality may also differ.

Previous works tackle the first challenge by designing modality-specific embedders and predictors to interface with the pretrained model from input to output. However, the second challenge have not yet been well addressed. Some approaches~\cite{metaTrans, onellm} argue that the large pretrained model can be served as universal encoder and thus freeze the pretrained model during finetuning. Other methods~\cite{fpt, orca, lift} finetune the pretrained model along with modality-specific components. Both line of works empirically show that the pretrained model can be transferred to other modalities. Still, the key problem of \textit{what knowledge from source modality is transferred via the pretrained model and how does it benefit the target modality} remains unsolved. For instance, ORCA~\cite{orca} observes that training the model from scratch on some target modality tasks is even better than the vanilla finetuning of the pretrained model, which indicates that the knowledge contained in the pretrained model may not improve target performance if it is not properly transferred.

In this work, we delve deeper into this second challenge of cross-modal transfer. We begin with experiments investigating how target modality finetuning affects the representation quality of the source modality data. It is observed that finetuning a pretrained Swin Transformer~\cite{swin} on some target modality tasks can help the Swin encoder to extract more discriminative features for images, while finetuning on other modalities impairs such ability. This empirical observation shows that there may exist aspects of knowledge, which we refer to as \textit{modality semantic knowledge}, that differ between modalities in different degree and affect the validity of cross-modal transfer. 

To specify such aspect of difference between modalities, we interpret the modality semantic knowledge as the conditional probability distribution $P(Y|X)$. We modify the conditional distribution of the source modality according to the tasks in target modality to make the two comparable. Consequently, we are able to formalize the \textit{modality knowledge discrepancy} in terms of the divergence between conditional distributions of source and target modality.\todo{sadly, this definition cannot be calculated for real tasks.} When the target conditional distribution is similar to the modified source conditional distribution, we say that the modality semantic knowledge is aligned and the source discriminative function learned by the pretrained model can be reused for the target modality. On the opposite, the modality semantic knowledge contradict each other and may not be mutually beneficial, which explains the observation in ORCA.

Our interpretation provides a new perspective towards understanding the effectiveness of the two-stage tuning pipeline proposed by previous cross-modal transfer works~\cite{lift, orca}: viewing the first stage as an implicit data transformation learning for target modality such that the conditional distribution on the transformed data are more aligned with source.
As a result, it enlightens us to directly learn a proper target embedding function ahead of finetuning, which helps minimize the knowledge misalignment.
To this end, we propose a new method, MoNA, that improves the cross-modal transfer with two-stage training. In the first stage, MoNA leverages meta learning to learn an optimal target embedder which, when served as an initialization along with the pretrained weights for the full finetuning, allows a maximum reuse of the source modality knowledge during full finetuning. In the second stage, using the learned target embedder as the starting point, we follow the vanilla finetuning approach and update all the parameters to adapt to the target task while maximally leveraging source knowledge.

We conduct extensive experiments on two cross-modal transfer benchmarks, NAS-Bench-360~\cite{nas-bench-360} and PDEBench~\cite{pdebench}, to validate our hypothesis and the effectiveness of our proposed methods. Both benchmarks focus on scientific problem related modalities, in which the training data scarcity is particularly acute. Comparisons of MoNA against previous methods are made, in which the results show that our method performs superior.


\section{Problem Formulation and Analysis}
\label{sec:problem}
In this section, we propose to test an assumption commonly made by previous cross-modal transfer approaches~\cite{fpt, orca, lift, metaTrans} that the pretrained model can serve as a universal encoder for different modalities. Our experiments lead to an intuitive conclusion that the knowledge gap between modalities are not the same, and thus the assumption should take the modality knowledge discrepancy into consideration.
\subsection{Introduction to basic notations and architecture}
We consider the knowledge transfer between source modality $\mathcal{M}^s$ and target modality $\mathcal{M}^t$. Data in source modality, such as vision or language, is easier and cheaper to obtain, and large pretrained models are publicly available. Instead, the target modality considered in this paper has insufficient data to pretrain its own large models. The two modalities differ in both input space and label space, i.e., $\mathcal{X}^s\neq\mathcal{X}^t$, $\mathcal{Y}^s\neq\mathcal{Y}^t$. Cross-modal transfer aims to leverage a source pretrained model, parameterized by $\boldsymbol{\theta}^{\mathcal{S}}$, to help a given target task with a small set of labeled data $\{\x_i^{t}, y_i^{t}\}_{i=1}^{n_t}$.

Following previous works~\cite{fpt,orca}, our model architecture $g_{\tha}$ includes an embedder $e(\cdot;\boldsymbol{\theta}_e)$, a transformer encoder $f(\cdot;\boldsymbol{\theta}_f)$ and a predictor $h(\cdot;\boldsymbol{\theta}_h)$, and the parameter of the full model is denoted as $\boldsymbol{\theta} = \{\boldsymbol{\theta}_e, \boldsymbol{\theta}_f, \boldsymbol{\theta}_h\}$. Particularly, the pretrained transformer has its own embedder and predictor, and thus we denote the pretrained weights of the source model as $\tha^{\Sm}_0 = \{\tha_{e_0}^\Sm, \tha_{f_0}^\Sm, \tha_{h_0}^\Sm\}$. 
\begin{figure*}[t]
  \centering
  \includegraphics[width=0.96\textwidth]{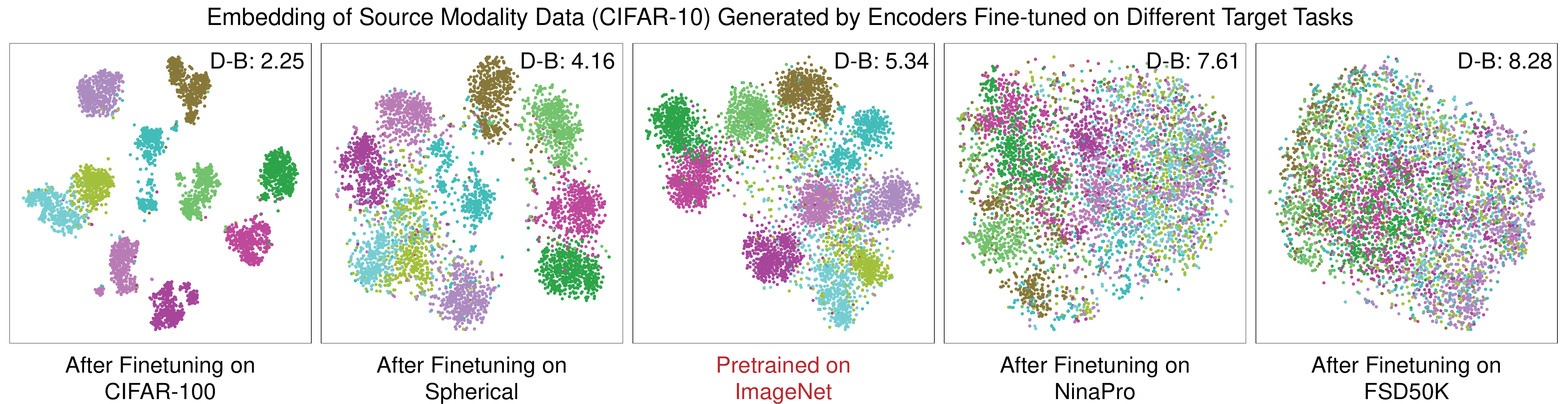}
  \vspace{-8pt}
  \caption{T-SNE visualization showing that the gap between different modalities are not the same. Figure in the middle depicts the embeddings of CIFAR-10 generated by an ImageNet pretrained Swin Transformer. The rest four figures are the embeddings of CIFAR-10 generated by the same model after being finetuned on different modalities. \textbf{None of these models are trained on CIFAR-10 directly}. Nevertheless, finetuning on CIFAR-100 and Spherical improve the visual representation from pretrained model while finetuning on NinaPro and FSD50K distort it. Davies-Bouldin indexes are shown at upper-right corner. Smaller index means better clustering.}
  \label{Fig:motive}
  \vspace{-14pt}
\end{figure*}

The embedder maps the input data into a shared input embedding space $\hat{\mathcal{X}}$, and the encoder extracts features from the embedded input. The predictor is a linear layer that maps the encoder output to the label space. For our target model $g_{\tha^\Tm}:\tha^\Tm = \{\tha^\Tm_e, \tha^\Tm_f, \tha^\Tm_h\}$, both embedder and the predictor are specifically re-designed to accommodate the input and label space of the target task, while we use $\tha_{f_0}^\Sm$ to initialize the encoder weight $\tha^\Tm_f$.

The flexibility of such architecture enables end-to-end training on the target task. \textbf{Vanilla finetuning} simply updates all the parameters of the target model by minimizing the task-specific loss on the given training dataset:
\vspace{-5pt}
\begin{equation}
    \tha^*_\Tm = \arg\min_{\tha_\Tm} \sum_{i=1}^{n_t} \ell\left(g_{\tha^\Tm}(\x_i^t), y_i^t\right),
    \label{eq:vanilla}
\end{equation}
\vspace{-5pt}
where $\ell$ is the task loss function such as cross-entropy.


Learning directly from target supervision in this way encourages the model to learn knowledge that help discriminate the target data. As the pretrained model already contains source discriminative knowledge, it is natural for cross-modal transfer to expect that source and target knowledge share similarities in some aspects so that the source knowledge can be reused to promote target learning. In the following, we 1) conduct experiments to show that this similarity depends on the modality, and 2) provide interpretation of modality knowledge and formalize the knowledge discrepancy.

\vspace{-5pt}
\subsection{Distortion of learned source modality knowledge}
We look for a quantitative way to compare the extent of knowledge reuse among various cross-modal transfer scenarios. In this section, we select image modality as knowledge source and choose four target tasks from different modalities. We include two tasks closely related to images: CIFAR-100~\cite{cifar100}, Spherical~\cite{spherical} that contains spherically projected images, and two tasks dissimilar to image modality:  NinaPro~\cite{ninapro} that represents hand gestures with electromyography signals, FSD50K~\cite{fsd} that contains audio clips of sound events. 
To be specific, we adopt Swin Transformer Base pretrained on ImageNet-22k as the source model and examine the properties of the model after finetuning it on different tasks.

Given that the comparison is conducted across distinct modalities, there lacks a general metric measuring the degree of knowledge reuse during transfer. Therefore, we turn to compare the distortion of source knowledge. Specifically, we would expect smaller distortion if more source knowledge is reused to solve target task, and vise versa.  
So we leverage the pretrained source model to extract the visual representation of CIFAR-10, a surrogate image dataset unseen by the model. Samples in this particular source dataset are denoted as $\{\x^s_i,y^s_i\}$ and their corresponding feature set $\{\boldsymbol{f}^s_i = f(e(\x^s_i;\tha_{e_0}^\Sm);\tha_{f_0}^\Sm)\}$. Then, we finetune the pretrained model on the four target tasks using Eq.~\eqref{eq:vanilla} respectively. After the finetuning process, we once again extract the representation of CIFAR-10 using the finetuned encoder and obtain $\{\boldsymbol{f}^s_i(\mathcal{M}_t) = f(e(\x^s_i;\tha_{e_0}^\Sm);\tha_f^\Tm, \mathcal{M}_t)\}$. Fig.~\ref{Fig:motive} shows the T-SNE visualization results of the five different sets of CIFAR-10 image features. 

The figure illustrates that encoders finetuned on CIFAR-100 or Spherical maintain or even improve their discriminability on image samples in CIFAR-10, whereas the encoders finetuned on NinaPro and FSD50K can no longer extract class discriminative features for images. Considering that finetuning on target modality makes the encoder focus on classifying target data and learning target discriminative function, what this observation indicates is that the knowledge required for discriminating samples in CIFAR-100 and Spherical are more aligned with that for CIFAR-10, compared to the latter two modalities. Such conclusion aligns with our intuition, since CIFAR-100 is vision dataset and Spherical is originated from natural images, whereas NinaPro and FSD50K are less relevant to images.

On the flip side, the results shows that CIFAR-100 and Spherical can better reuse the source knowledge in the pretrained encoder for task solving while NinaPro and FSD50K require the encoder to make greater adjustments in order to adapt to target tasks.

To investigate the source knowledge reuse (or distortion) during cross-modal transfer more quantitatively, we use linear probing on CIFAR-10 to evaluate the quality of extracted representations with encoders finetuned 1) on different target modalities, 2) with different epochs, and 3) with different transfer methods. Additional to vanilla finetuning, we consider the following two baselines:

\begin{itemize}
  \item \textbf{ORCA}~\cite{orca} adds an embedder training stage before finetuning. This first stage only updates the target embedder parameters $\tha_e^\Tm$ to minimize the Optimal Transport Dataset Distance~\cite{otdd} between source and target embeddings in the shared input space $\hat{\mathcal{X}}$. 
  \item We propose another baseline modified from previous works~\cite{theory_ftvslp,metaTrans}, Embedder warmup (\textbf{Emb}), which is also a two-stage training method. The first stage solely updates the target embedder using the same task loss as vanilla finetuning while keeping the rest of the network frozen. The second stage finetunes the full network.
\end{itemize}

\begin{figure}[htbp]
  \centering
  \includegraphics[width=0.48\textwidth]{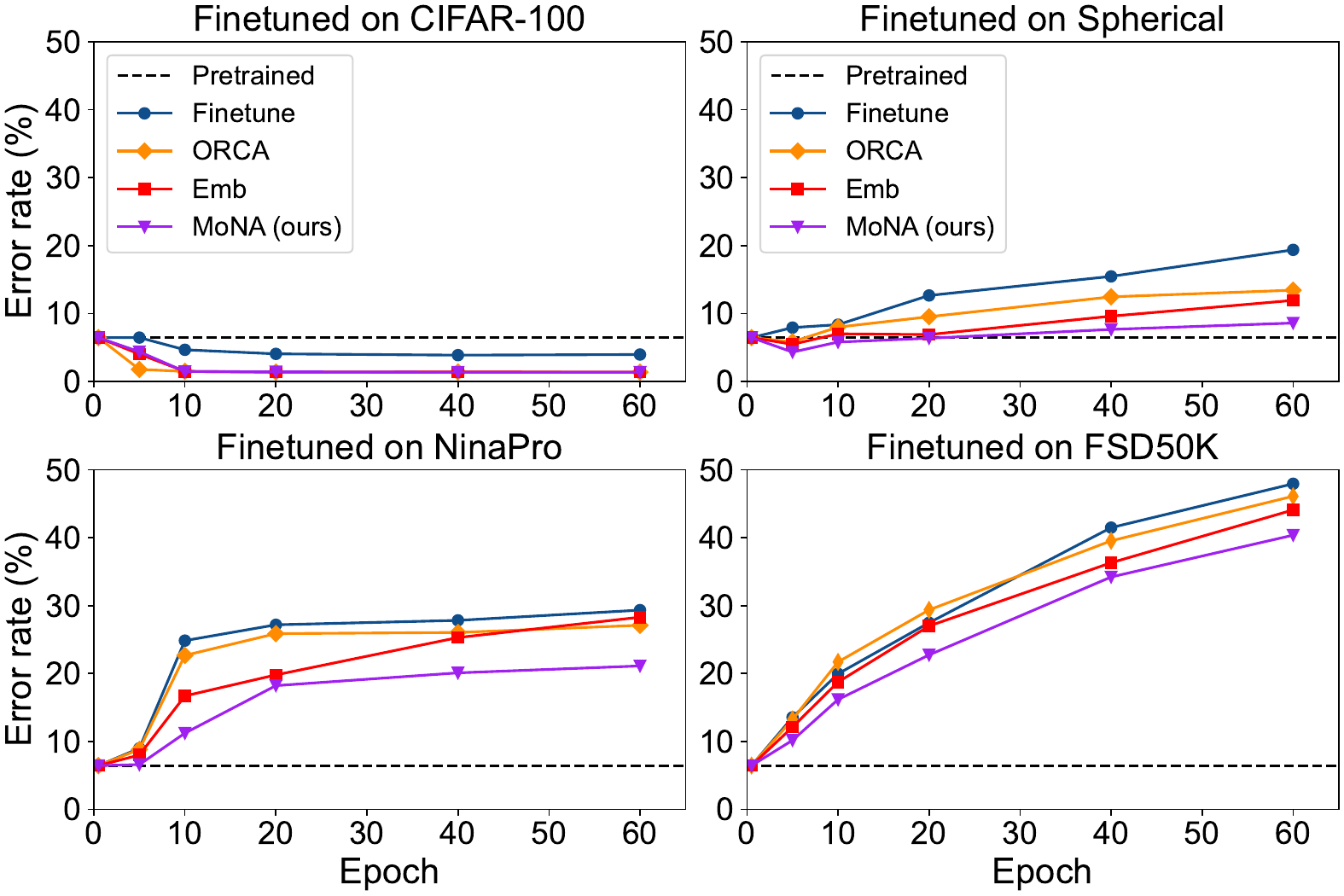}
  \vspace{-18pt}
  \caption{Linear probing results on CIFAR-10 using representations extracted by vision encoders finetuned on four different modalities and with different finetuning methods. ``Pretrained'' refers to the baseline that directly uses pretrained vision encoder.}
  \label{Fig:sourcelp}
  \vspace{-14pt}
\end{figure}

Fig.~\ref{Fig:sourcelp} shows the \textit{error} rate of linear probing, where the dash line shows the linear probing results on pretrained encoder as a reference. Note that all these results are error rates on CIFAR-10 dataset that reflects the extent to which the model retains the source modality knowledge. The performance comparison on target modalities is not what we concern right now and can be found later in table.~\ref{tab:variants}. From experiments we observe that modality has the greatest effect on linear probing results. Finetuning on FSD50K significantly distorts the encoder and impairs its discriminability on image data. We also notice that tuning on target dataset for more epochs leads to larger distortion of source knowledge on all target modalities except for image modality (CIFAR-100). These observation leads to the conclusion that knowledge for discriminating samples in different modalities differ in varying degrees, which we refer to as the misalignment of modality semantic knowledge. We argue that a large discrepancy may hinder the effectiveness of cross-modal transfer, and thus the assumption that source modality pretraining is beneficial to target modality should depends on the discrepancy.

We make additional observations about the source knowledge preservation effect of two-stage training methods. We observe that both ORCA and Emb achieves lower source error compared to vanilla finetuning, and Emb performs better than ORCA. This suggests that the target embedder trained in their first stage implicitly learns a mapping from $\mathcal{X}^t$ to $\hat{\mathcal{X}}$ that mitigates the knowledge misalignment between target and source, and thus reduces the model distortion during its adaptation towards target tasks.

The above experiments motivate us to formalize the discrepancy of modality semantic knowledge, and to propose an improved objective for training target embedders that reduce such discrepancy.
\subsection{Modality semantic knowledge discrepancy}
We consider representing the semantic knowledge within a modality using the conditional distribution $P(Y|X)$, which describes the relationship between raw data space and semantic space of the modality.
This is because for neural networks, acquiring the semantic knowledge means learning a mapping from data space to semantic space that resembles the true conditional distribution.

However, to measure the degree of alignment or ``similarity'' of such knowledge between two modalities is quite challenging. The difficulty lies in the fact that both the data space $\mathcal{X}$ and the label space $\mathcal{Y}$ are different and even non-overlapping across modalities. 

Therefore, we need to modify the conditional distribution to make it comparable across modalities. Modifying the input space is rather easy, as we can simply embed the inputs into a shared space $\hat{\mathcal{X}}$ using modality-specific embedders. However, modifying the label space is more sophisticated.

Considering that source modalities, like vision and language, having large pretrained models are both rich in semantics, we make the following assumption:
\textit{The cardinality of source modality label space is larger than the cardinality of the label space of the target modality, i.e., $|\mathcal{Y}^s| \ge |\mathcal{Y}^t|$.}

This assumption is easily satisfied in practice. For example, vision transformers trained on ImageNet learns a discriminative function of one thousand categories whereas only four classes are considered in an electrocardiogram classification task. With the assumption, we can select a subset of the source modality label space $\mathcal{Y}^s_{\mathcal{B}} \subset \mathcal{Y}^s$ such that $|\mathcal{Y}^s_{\mathcal{B}}| = |\mathcal{Y}^t|$. We further introduce a category permutation $\pi(\cdot)$ that adjusts the order of source classes. To this end, we can define a new label space of the source modality, namely the source subset after permutation $\mathcal{Y}^s_{\pi,\mathcal{B}} \triangleq \pi(\mathcal{Y}^s_{\mathcal{B}})$. By measuring the discrepancy between the modified conditional distributions $P(Y^s_{\pi,\mathcal{B}}|\hat{X})$ and $P(Y^t|\hat{X})$, we can formalize the degree of alignment of the modality semantic knowledge as follows:


\begin{definition}\label{def:diff}
    (Modality semantic knowledge discrepancy). \textit{Given the source modality $\mathcal{M}^s$ and the target modality $\mathcal{M}^t$ satisfying the assumption, let $\hat{mathcal{X}}$ be the shared input space generated from raw data spaces by modality-specific embedders, and let $P(Y^s|\hat{X})$, $P(Y^t|\hat{X})$ be the conditional distribution for the source and target modality.
    Then, the modality semantic knowledge discrepancy between the two modalities is $$D(\mathcal{M}^s, \mathcal{M}^t) = \inf_{\pi,\mathcal{B}} d(P(Y^s_{\pi,\mathcal{B}}|\hat{X}), P(Y^t|\hat{X})),$$
    where $d(\cdot,\cdot)$ is an arbitrary discrepancy measure between two conditional distributions.}
\end{definition}
The definition basically says that, if we can find an optimal subset within source semantics that, with a proper one-to-one matching between source semantics and target semantics, shares similar conditional distribution with target modality, then the knowledge discrepancy is considered to be small. The source model should be able to correctly distinguish target samples like the way it discriminates source samples within the subset.

\begin{figure}[htbp]
  \centering
  \vspace{-15pt}
  \includegraphics[width=0.45\textwidth]{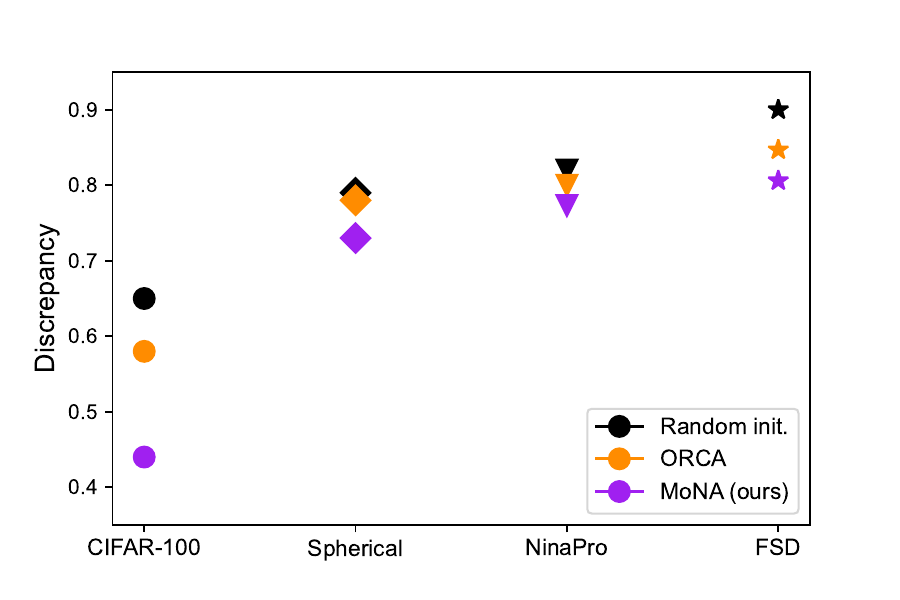}
  \vspace{-20pt}
  \caption{Modality knowledge discrepancy between image modality and four target modalities. Computation uses approximation.}
  \label{Fig:disc}
\end{figure}
With the definition, we calculate the modality semantic knowledge discrepancy between image modality and the four target tasks using an extreme approximating algorithm. Here we only demonstrate the results in Fig.~\ref{Fig:disc} while leaving the implementation details in supplementary. Our calculation aligns with previous observations, showing that modalities do have different degree of knowledge discrepancy, and FSD50K is the most dissimilar modality from image modality among the four.






\begin{figure*}[t]
    \centering
	\vspace{-5pt}
    \includegraphics[width=0.99\textwidth]{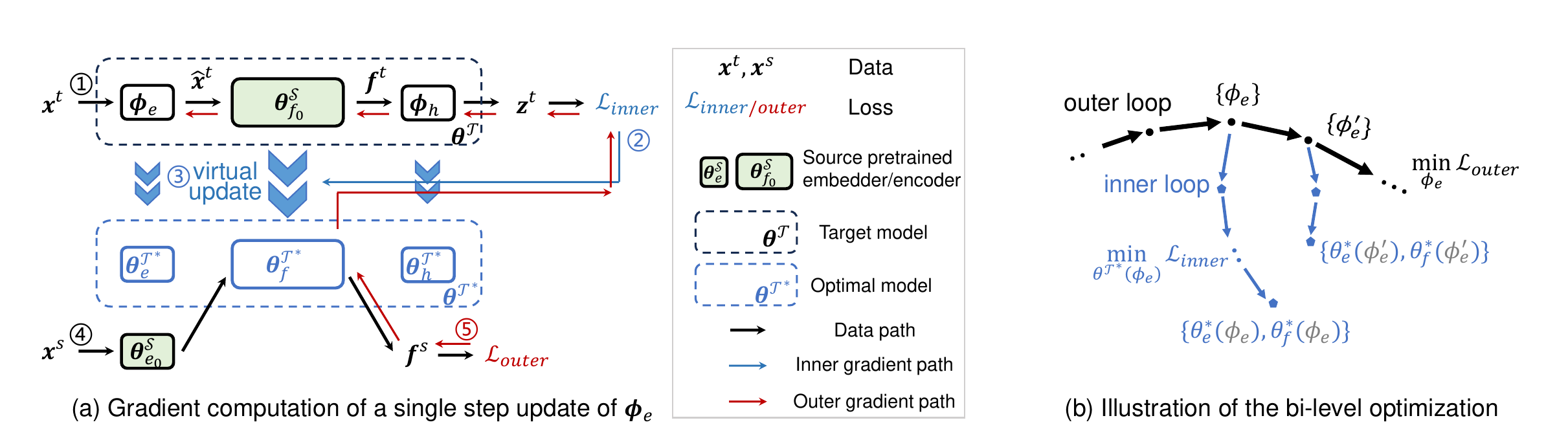}
    \vspace{-15pt}
    \caption{The framework of our proposed method. (a) illustrates a single update step of the embedder $\phi_e$ in the first stage of MoNA. The target data is forward propagated first to compute the inner-loop loss $\Lm_{inner}$, and the gradient is backpropagated to virtually update the full target model. Then, the updated encoder $\boldsymbol{\theta}^{\mathcal{T}^*}_f$ receives source data embeddings from pretrained source embedder $\boldsymbol{\theta}_{e_0}^{\mathcal{S}}$, and the outer-loop loss is computed using source features. Finally, the outer-loop gradient is used to update the embedder while the virtually updated model is discarded. (b) illustrated the bi-level optimization where the outer-loop updates $\boldsymbol{\phi}_e$ according to inner-loop results.}
    \label{Fig:framework}
\end{figure*}

\section{Modality Knowledge Alignment}
Discovering that the modality knowledge may not be well aligned and its consequence of insufficient source knowledge reuse, we propose a new method that improves the modality knowledge alignment and the effectiveness of cross-modality transfer.  

\vspace{-8pt}
\subsection{Embedder Warmup}
In the previous experiments we find that Embedder warmup, in spite of its simplicity in training objective, preserves source knowledge better than other methods. Correspondingly, we turn to examine its performance on target modalities. Table~\ref{tab:variants} shows that Emb likewise surpasses its counterparts. We argue that during the embedder warmup, in order to minimize the task loss, the embedder are forced explicitly to project target original inputs into embeddings that are distinguishable by the source model, which is frozen and extract features according to the source knowledge. 

Combined with our previous analysis, we hypothesize that the key to effective transfer is to learn a target embedding function $e^\Tm: \mathcal{X}\to \hat{\mathcal{X}}$ that makes the target conditional distribution $P(Y^t|\hat{X})$ more aligned with the source knowledge. Consequently, we propose to train the target embedder solely using objective in the next section to learn such embedding function ahead of the full finetuning process. 



\vspace{-8pt}
\subsection{Learning to Align Modality Knowledge}
Since we cannot estimate the target conditional probability without training a model, adopting the modality knowledge divergence directly as a objective for optimization is difficult. As an alternative, we propose to leverage meta learning pipeline to simulate the process in Fig.~\ref{Fig:sourcelp} and optimize the representation quality of source data after finetuning. Specifically, an ideal target embedder aligns the modality knowledge, allowing the encoder to retain its discriminability on image data during target finetuning. Therefore, if we use a source dataset to evaluate the finetuned encoder that is initialized by this ideal target embedder, we would obtain minimal error on the source data.

Such process is a standard bi-level optimization problem widely studied in meta-learning~\cite{maml}. Particularly in our case, the outer-loop updates the target embedder based on the outer-loop loss, which is computed using target encoder after inner-loop optimization. Fig.~\ref{Fig:framework}(a) illustrates a single step update of the embedder parameter $\boldsymbol{\phi}_e$ in the outer-loop during meta learning, and Fig.~\ref{Fig:framework}(b) shows the process of bi-level optimization. 

More specifically, the inner-loop is the optimization of the model on target dataset, subjected to the condition that the target embedder is initialized by $\boldsymbol{\phi}_e$, which is
\begin{equation}
  \tha^{\Tm^*}(\boldsymbol{\phi}_e) = \arg\min_{\tha^\Tm} \Lm_{inner}(\x^t, y^t; \boldsymbol{\phi}_e),
\end{equation}\label{eq:inner}
where $\Lm_{inner}$ is the same loss as in Eq.~\eqref{eq:vanilla}, and 
\begin{equation}
  \tha_{e_t}^\Tm= \tha_{e_{t-1}}^\Tm - \alpha \nabla \Lm_{inner}, \quad \tha_{e_0}^\Tm=\boldsymbol{\phi}_e.
\end{equation}\label{eq:inner_update}
This inner-loop optimization simulates the full finetuning process in the second stage, and returns an encoder that is already adapted to target modality. Note that the whole optimal target model in the inner-loop depends on the initialization of the target embedder. Therefore we have $\tha^{\Tm^*}(\boldsymbol{\phi}_e) = \{ \tha^{\Tm^*}_e(\boldsymbol{\phi}_e),\tha^{\Tm^*}_f(\boldsymbol{\phi}_e),\tha^{\Tm^*}_h(\boldsymbol{\phi}_e) \}$.

The outer-loop is an optimization problem with respect to the target embedder. Our goal is to find optimal embedder parameters ${\phi}_e^*$ such that the resulting optimal target encoder $\tha^{\Tm^*}_f(\boldsymbol{\phi}_e^*)$ generates high quality representations of source data. To calculate the loss, we leverage a small labeled dataset $\{\x_i^s, y_i^s\}$ in source modality as a surrogate and compute their features $\{\boldsymbol{f}^s_i = f(e(\x^s_i;\tha_{e_0}^\Sm);\tha_f^{\Tm^*}(\boldsymbol{\phi}_e))\}$. Then we normalize these features onto a unit sphere and measure the alignment and uniformity of the source features~\cite{alignment_uniform}. In particular, the alignment loss measures whether features from the same class are close, and the uniformity loss measures whether features from different classes are evenly distributed on the sphere.

\begin{algorithm}[t!]
  \caption{MoNA: Modality Knowledge Alignment}\label{alg:mona}
  \begin{algorithmic}
  \STATE {\bfseries Input:}
      Source pretrained model $g_{\tha^\Sm_0}$; Learning rate $\alpha,\beta$; Maximum iterations $I_1, I_2$.
  \STATE {\bfseries Output:}
      Model for the target task: $g_{\tha^\Tm}$.
  \STATE {\bfseries Stage 1: Target embedder training}
  \FOR{$iter=1,2,\cdots,I_1$}
  \STATE Initialize the target model $g_{\tha^\Tm}$ with $\boldsymbol{\phi}_e$ and ${\tha^\Sm_{f_0}}$.
  \STATE Virtually update: ${\tha^{\Tm^*}}={\tha^\Tm}-\alpha\nabla_{\tha^\Tm}\Lm_{inner}$.
  \STATE Compute source features $\{\boldsymbol{f}^s_i\}$ with $\tha_{e_0}^\Sm$ and $\tha_f^{\Tm^*}$.
  \STATE Obtain outer-loop loss using Eq.~\eqref{eq:au}.
  \STATE Update target embedder: $\boldsymbol{\phi_e}\leftarrow\boldsymbol{\phi_e}-\beta\nabla_{\phi_e}\Lm_{outer}^{'}$.
  \ENDFOR
  \STATE {\bfseries Stage 2: Full finetuning}
  \FOR{$iter=1,2,\cdots,I_2$}
  \STATE Initialize the target model $g_{\tha^\Tm}$ with $\boldsymbol{\phi}_e$ and ${\tha^\Sm_{f_0}}$.
  \STATE Update target model towards Eq.~\eqref{eq:vanilla}.
  \ENDFOR
  \end{algorithmic}
\end{algorithm}
Our outer-loop objective that measures the source discriminability of the induced encoder takes the following form:
\vspace{-10pt}
\begin{equation}\label{eq:au}
\begin{aligned}
 \Lm_{outer} &= \Lm_{align} + \Lm_{uniform} \\
 &= - \mathop{\mathbb{E}}\limits_{i,j:y^s_i=y^s_j} [\vert\vert \boldsymbol{f}^s_i-\boldsymbol{f}^s_j \vert\vert_2^2] - \log \mathop{\mathbb{E}}\limits_{i,j} \left[e^{-2\vert\vert \boldsymbol{f}^s_i-\boldsymbol{f}^s_j \vert\vert_2^2}\right].
\end{aligned}
\end{equation}
Notably, the source knowledge cannot be well-preserved at the beginning of the embedder training. To prevent the embedder from overly focusing on source modality, and also to keep the optimization process stable, we strike a balance between source and target knowledge learning by jointly minimizing the two objectives with a trade-off parameter $\lambda$:
\vspace{-8pt}
\begin{equation}\label{eq:outer_prime}
  \Lm_{outer}^{'} = \lambda \Lm_{outer} + \Lm_{inner}.
\end{equation}
In practice, we adopt single step update in the inner loop, a simplification that will be discussed in the analytical experiment section. This enables us to reuse the loss $\Lm_{inner}$ calculated during the inner-loop virtual update to compute this combined objective $\Lm_{outer}^{'}$ efficiently.
To this end, our proposed MoNA updates the target embedder in the first stage using:
\begin{equation}
  \boldsymbol{\phi}_e^* = \arg\min_{\boldsymbol{\phi}_e} \Lm_{outer}^{'}.
\end{equation}\label{eq:outer}
With the modality knowledge being better aligned, MoNA conducts vanilla finetuning in the second stage. The complete algorithm of MoNA is demonstrated in Alg.~\ref{alg:mona}.

\vspace{-8pt}
\section{Experiments}
\vspace{-5pt}
\begin{table*}[t]
  \centering
  \vspace{-10pt}
  \caption{Prediction errors ($\downarrow$) on ten tasks of NAS-Bench-360.}
  \resizebox{0.996\textwidth}{!}{%
    \begin{tabular}{lccccccccccccccccccccc}
    \toprule
    {\multirow{2}[2]{*}{}} & {CIFAR-100} & {Spherical} & {Darcy Flow} & {PSICOV} & {Cosmic} & {NinaPro} & {FSD50K} & {ECG} & {Satellite} & {DeepSEA} \\
    {} & {0-1 error (\%)} & {0-1 error (\%)} & {relative $l_2$} & {MAE8} & {1-AUROC} & {0-1 error (\%)} & {1-mAP} & {1-F1 score} & {0-1 error (\%)} & {1-AUROC} \\
    \midrule
    {Hand-designed} & {19.39} & {67.41} & {8.00E-3} & {3.35} & {0.127} & {8.73} & {0.62} & {0.28} & {19.8} & {0.3} \\
    {NAS-Bench-360} & {23.39} & {48.23} & {2.60E-3} & {2.94} & {0.229} & {7.34} & {0.6} & {0.34} & {12.51} & {0.32} \\
    {DASH} & {24.37} & {71.28} & {7.90E-3} & {3.3} & {0.19} & {\textbf{6.60}} & {0.6} & {0.32} & {12.28} & {\textbf{0.28}} \\
    {Perceiver IO} & {70.04} & {82.57} & {2.40E-2} & {8.06} & {0.485} & {22.22} & {0.72} & {0.66} & {15.93} & {0.38} \\
    \midrule
    {FPT} & {10.11} & {76.38} & {2.10E-2} & {4.66} & {0.233} & {15.69} & {0.67} & {0.5} & {20.83} & {0.37} \\
    {ORCA} & {6.53} & {29.85} & {7.28E-3} & {1.91} & {0.152} & {7.54} & {0.56} & {0.28} & {11.59} & {0.29} \\
    \rowcolor{gray!30} {\textbf{MoNA}} & {\textbf{6.48}} & {\textbf{27.13}} & {\textbf{6.80E-3}} & {\textbf{0.99}} & {\textbf{0.121}} & {7.28} & {\textbf{0.55}} & {\textbf{0.27}} & {\textbf{11.13}} & {\textbf{0.28}} \\
    \bottomrule
    \end{tabular}%
  }
  \vspace{-15pt}
  \label{tab:nasbench}%
\end{table*}%
\begin{table*}[t!]
  \centering
  \caption{Comparing various cross-modal transfer methods on NAS-Bench-360.}
  \resizebox{0.996\textwidth}{!}{%
    \begin{tabular}{lccccccccccccccccccccc}
    \toprule
    {{\multirow{2}[2]{*}{}}} & {CIFAR-100} & {Spherical} & {Darcy Flow} & {PSICOV} & {Cosmic} & {NinaPro} & {FSD50K} & {ECG} & {Satellite} & {DeepSEA} \\
    {} & {0-1 error (\%)} & {0-1 error (\%)} & {relative $l_2$} & {MAE8} & {1-AUROC} & {0-1 error (\%)} & {1-mAP} & {1-F1 score} & {0-1 error (\%)} & {1-AUROC} \\
    \midrule
    {Train-from-scratch} & 50.87 & 76.67 & 8.00E-2 & 5.09 & 0.50 & 9.96 & 0.75 & 0.42 & 12.38 & 0.39 \\
    \midrule
    {Finetuning} & 7.67 & 55.26  &  7.34E-3 & 1.92 & 0.17 & 8.35 & 0.63 & 0.44 & 13.86 & 0.51\\
    {Finetuning++} & 6.60 & 33.17 & 7.50E-3 & 1.91 & 0.168 & 8.00 & 0.63 & 0.35 & 12.73 & 0.38\\             
    {Frozen-encoder} & 10.02 & 59.62 & 6.84E-3  & 3.43 & 0.481 & 35.20& 0.72 & 0.37 & 19.71 & 0.36\\
    \midrule
    {ORCA} & {6.53} & {29.85} & {7.28E-3} & {1.91} & {0.152} & {7.54} & {0.56} & {0.28} & {11.59} & {0.29} \\
    {Emb} & 6.52 & 28.76 & 7.50E-3 & 1.35 & 0.139 & 7.74 & 0.56 & 0.28 & 11.40 & 0.29 \\
    \rowcolor{gray!30} {\textbf{MoNA}} & {\textbf{6.48}} & {\textbf{27.13}} & {\textbf{6.80E-3}} & {\textbf{0.99}} & {\textbf{0.121}} & {7.28} & {\textbf{0.55}} & {\textbf{0.27}} & {\textbf{11.13}} & {\textbf{0.28}} \\
    \bottomrule
    \end{tabular}%
  }
  \vspace{-12pt}
  \label{tab:variants}%
\end{table*}%
In this section, we show experiments conducted on two cross-modal benchmarks. We follow the test protocol in ORCA~\cite{orca} and evaluate MoNA on NAS-Bench-360~\cite{nas-bench-360} and PDEBench~\cite{pdebench}. NAS-Bench-360 is a comprehensive benchmark that contains diverse tasks from ten different modalities. PDEBench covers a wide range of partial differential equations (PDEs) including challenging physical problems.

Each benchmark involves both tasks with 1D and 2D inputs. Following previous works, we adopt pretrained language model RoBERTa~\cite{roberta} and pretrained vision model Swin Transformer~\cite{swin} for 1D and 2D tasks respectively. Following ORCA, We use CoNLL-2003 and CIFAR-10 as the source modality datasets to compute the outer-loop meta loss. Hyper-parameters for each task are in supplementary.

\vspace{-5pt}
\subsection{Results on NAS-Bench-360}
\vspace{-5pt}
Several experiments are conducted on this benchmark. The benchmark involves seven tasks with 2D inputs and three tasks with 1D inputs. Table~\ref{tab:nasbench} shows the results comparison using the \textit{full} training set in each modality. We compare different kinds of baselines including training hand-designed and NAS-based architectures (NAS-Bench-360, DASH~\cite{dash}) solely on target modalities, general purpose networks Perceiver IO~\cite{perceiverIO}, and cross-modal transfer methods FPT~\cite{fpt} and ORCA. 
We observe that MoNA achieves top performance on nine out of ten tasks, and surpasses previous cross-modal transfer methods on all tasks.


We also compares several variants of cross-modal transfer. In single stage methods, finetuning refers to the vanilla finetuning in Eq.~\eqref{eq:vanilla}. 
Finetuning++ initializes the target embedder using source embedder weights under certain modifications to map the dimension. 
Frozen encoder resembles to previous works~\cite{metaTrans} and it keeps the pretrained weight frozen during finetuning. In two-stage methods where a warmup stage is conducted before finetuning, we consider ORCA, Emb and MoNA. These experiments serves as an ablation study of our method and their results is reported in table~\ref{tab:variants}.

\begin{table}[t]
  \centering
  \Huge
  \caption{Performance Comparison ($\downarrow$) with Different Backbones.}
  \resizebox{0.485\textwidth}{!}{%
    \begin{tabular}{lccccc}
    \toprule
    {{\multirow{2}[1]{*}{Method}}} & {{\multirow{2}[1]{*}{Pretrained Model (Size)}}} & {{\multirow{2}[1]{*}{Pretrained Dataset}}} & Spherical & NinaPro & FSD \\
     && & 0-1 error (\%) & 0-1 error (\%) &{1-mAP} \\
     \midrule
    Finetuning & ViT-Base (86M) & IN-22K & 47.24 & 15.63 & 0.74 \\
    ORCA & ViT-Base (86M) & IN-22K & 36.52 & 8.78 & 0.63 \\
    \rowcolor{gray!30} 
    \textbf{MoNA} & ViT-Base (86M) & IN-22K & \textbf{33.34} & \textbf{8.00} &\textbf{0.62} \\
    \midrule
    Finetuning & CLIP ViT-B/16 (86M) & WIT-400M & 57.47 & 13.81 & 0.77 \\
    ORCA & CLIP ViT-B/16 (86M) & WIT-400M & 43.12 & 8.50 & 0.69 \\
    \rowcolor{gray!30} 
    \textbf{MoNA} & CLIP ViT-B/16 (86M) & WIT-400M & \textbf{41.35} & \textbf{7.59} &\textbf{0.67} \\
    \midrule
    Finetuning & Swin-Base (88M) & IN-22K & 55.26 & 8.35 & 0.63 \\
    ORCA & Swin-Base (88M) & IN-22K & 29.85 & 7.54 & 0.56 \\
    \rowcolor{gray!30} 
    \textbf{MoNA} & Swin-Base (88M) & IN-22K & \textbf{27.13} & \textbf{7.28} &\textbf{0.55} \\
    \bottomrule
    \end{tabular}%
  }
  \vspace{-14pt}
  \label{tab:diff_models}%
\end{table}%

\todo{column width}
\begin{table*}[htbp]
    \centering
    \vspace{-10pt}
    \caption{Normalized Root Mean Squared Errors (nRMSEs, $\downarrow$) on 8 PDEBench tasks.}
    \resizebox{0.996\textwidth}{!}{%
      \begin{tabular}{ccccccccccccccccc}
      \toprule
      \multirow{2}[2]{*}{} & {Advection} & {Burgers} & {Diffusion-Reaction} & {Diffusion-Sorption} & {Navier-Stokes} & {Darcy-Flow} & {Shallow-Water} & {Diffusion-Reaction} \\
            & {1D} & {1D} & {1D} & {1D} & {1D} & {2D} & {2D} & {2D} \\
      \midrule
      PINN  & {0.67} & {0.36} & {0.006} & {0.15} & {0.72} & {0.18} & {0.083} & {0.84} \\
      FNO   & {0.011} & {\textbf{0.0031}} & {\textbf{0.0014}} & {0.0017} & {0.068} & {0.22} & {\textbf{0.0044}} & {\textbf{0.12}} \\
      U-Net & {1.1} & {0.99} & {0.08} & {0.22} & {--} & {--} & {0.017} & {1.6} \\
      \midrule
      ORCA  & {0.0098} & {0.0120} & {0.0030} & {\textbf{0.0016}} & {0.062} & {0.081} & {0.0060} & {0.820} \\
      \rowcolor{gray!30} \textbf{MoNA} & {\textbf{0.0088}} & {0.0114} & {0.0028} & {\textbf{0.0016}} & \textbf{0.054} & {\textbf{0.079}} & {0.0057} & {0.818} \\
      \bottomrule
      \end{tabular}%
    }
    \vspace{-10pt}
    \label{tab:pdebench}%
  \end{table*}%

The results lead to following conclusions: 1) finetuning the encoder helps the pretrained model adapt to target modality and performances better than frozen encoder. 2) Two-stage methods are generally superior than single stage methods, indicating that a proper target embedding function leads to better knowledge transfer. 3) Combined with the results on source knowledge preservation experiment in Fig.~\ref{Fig:sourcelp}, we see that methods that better align modality knowledge achieves higher cross-modal transfer performance. 

To show that MoNA achieves consistent performance improvement across different pretrained models, we conduct experiments on other two vision backbones, which are ViT-Base~\cite{vit} pretrained on ImageNet-22K and CLIP ViT-Base/16~\cite{clip} pretrained on WIT-400M. The results in table~\ref{tab:diff_models} show that, although the performance on target modalities varies due to the different capability of the pretrained models, MoNA consistently improves knowledge transfer and is superior to the State-of-the-Art method on different pretrained models.

\subsection{Results on PDEBench}

The benchmark includes eight PDEs with 1D/2D inputs, which we address similarly using language and vision pretrained models. We consider three baselines in the original paper, namely U-Net~\cite{u-net}, PINN~\cite{pinn}, FNO~\cite{fno}, where the last two methods are specifically designed for PDEs. We also compare ORCA as the cross-modal transfer baseline. The result is shown in table~\ref{tab:pdebench}.

We observe that MoNA achieves state-of-the-art on four out of eight tasks on PDEBench. Notably, it outperforms ORCA on seven tasks with significant improvements on Advection Equation and Navier-Stokes' Equation, and it achieves competitive results with specialized method FNO.

\vspace{-5pt}
\subsection{Results on Several Other Tasks}
\vspace{-5pt}
To further demonstrate the scalability of MoNA, we conduct experiments on more datasets and modalities, including AudioSet~\cite{audioset} and ESC50 in audio modality, and UCF101~\cite{ucf101} in video modality. 

As shown in table~\ref{tab:other_tasks}, MoNA surpasses all baselines including ORCA. These results validate the scalability of our method, indicating that MoNA is a general cross-modality transfer method that can be applied on a wide range of tasks.

\vspace{-5pt}
\subsection{Analytical Experiments}
\vspace{-5pt}
\begin{table}[t]
  \centering
  \tiny
  \caption{Prediction errors (↓) on three classic benchmarks from different modalities.}
  \resizebox{0.485\textwidth}{!}{%
    \begin{tabular}{lccc}
    \toprule
     & AudioSet-20k & ESC50 & UCF101  \\
     & 1-mAP & 0-1 error (\%) & 0-1 error (\%) \\
    \midrule
    From scratch & 0.634 & 30.00 & 57.62  \\
    Finetuning & 0.541 & 27.50 & 26.51 \\
    ORCA & 0.538 & 12.75 & 16.86\\
    \rowcolor{gray!30}
    \textbf{MoNA} & \textbf{0.523} & \textbf{9.64} & \textbf{13.45} \\

    \bottomrule
    \end{tabular}%
  }
  \vspace{-14pt}
  \label{tab:other_tasks}%
\end{table}%

\begin{table}[t]
  \centering
  \Huge
  \caption{Ablation Studies on Loss Design.}
  \resizebox{0.485\textwidth}{!}{%
    \begin{tabular}{lcccc}
    \toprule
    Loss Ablations & CIFAR-100 & Spherical & NinaPro & FSD50K \\
    \midrule
    MoNA w/o $\Lm_{inner}$ in Eq.~\eqref{eq:outer_prime} & 8.00& 28.76& 7.74 & 0.58 \\
    \rowcolor{gray!30}
    \textbf{MoNA} & \textbf{6.48} & \textbf{27.13} & \textbf{7.28} & \textbf{0.55} \\
    Contrastive Loss for $\Lm_{outer}$ & 6.51 & 27.90 & 7.44 & 0.55 \\
    Clustering Metric for $\Lm_{outer}$ & 7.09 & 28.02 & 8.19 & 0.56\\
    \bottomrule
    \end{tabular}%
  }
  \vspace{-18pt}
  \label{tab:ablation}%
\end{table}%

\textbf{Ablation Studies}. To validate the effectiveness of our design, we conduct several ablation studies on four tasks in NAS-Bench-360. The first ablation study investigates the effect of different training objectives as shown in table~\ref{tab:ablation}. We begin with excluding the inner-loop loss $\mathcal{L}_{inner}$ from the total objective of the outer-loop in Eq.~\eqref{eq:outer_prime}. Comparing to MoNA, the performance drop on all four tasks shows that $\Lm_{inner}$ is crucial to ensure the adaptation of the embedder.

\begin{figure}[t]
  \centering
  \vspace{5pt}
  \includegraphics[width=0.48\textwidth]{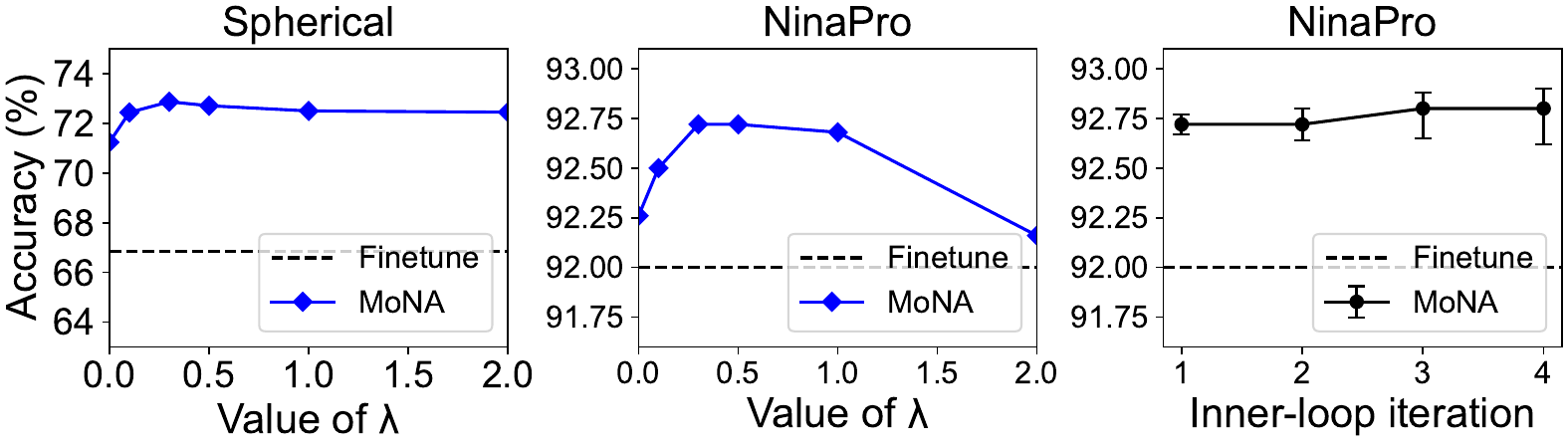}
  \vspace{-22pt}
  \caption{Analytical experiments. Accuracies with varying value of trade-off $\lambda$ (left two), and different inner-loop iterations (right).}
  \label{Fig:exp}
  \vspace{-12pt}
\end{figure}

\begin{table}[t]
  \centering
  \LARGE
  \caption{Comparison on Training Time in GPU Hour.}
  \resizebox{0.485\textwidth}{!}{%
    \begin{tabular}{lcccc}
    \toprule
      & CIFAR-100 & Spherical & NinaPro & FSD50K \\
    \midrule
    ORCA & 11.57 & 12.40 & 1.42 & 16.20 \\
    MoNA & 12.15 & 12.32 & 1.15 & 17.74 \\
    \midrule
    Performance Gain (relative) & +0.7\% & +9.1\% & +3.4\% & +1.8\% \\
    \bottomrule
    \end{tabular}%
  }
  \vspace{-14pt}
  \label{tab:time}%
\end{table}%

We move on to replace the alignment and uniformity loss (i.e., $\Lm_{outer}$) by two variants. The contrastive loss refers to the supervised contrastive loss~\cite{supCon}, and the clustering metric we adopt is the Davies-Bouldin index. The results in table~\ref{tab:ablation} shows that contrastive loss achieves slightly worse performance than MoNA, while the clustering metric leads to much worse results. We hypothesize the reason is that the DB index becomes less informative when the feature dimension is high. 

The ablations on different training strategies can be found in table~\ref{tab:variants}. Comparing two-stage MoNA against one-stage finetuning method, we find that the knowledge alignment stage brings significant improvement.


\textbf{MoNA reduces modality knowledge misalignment}. We empirically validate that MoNA reduces the discrepancy between source modality and four target tasks (Fig.~\ref{Fig:disc}), and by which better retains source knowledge during finetuning than other methods (Fig.~\ref{Fig:sourcelp}). Our empirical results align with our hypothesis that reducing knowledge misalignment between modalities leads to more effective transfer and higher performance on target tasks.

\textbf{Parameter sensitivity of $\lambda$}. This parameter balances the trade-off between adapting to target knowledge and preserving source knowledge. Fig.~\ref{Fig:exp}(a)(b) shows results on Spherical and NinaPro tasks with varying $\lambda$. We observe that optimal value lies in the range of $[0.3,0.5]$, whereas much higher or lower values lead to performance drop.

\textbf{Inner-loop update steps}. We take different steps of gradient descent in the inner-loop and show the results in Fig.~\ref{Fig:exp}(c). We observe that increasing the inner-loop steps slightly improves the performance. However, it also causes an increasing accuracy variation as well as computational cost. 
Therefore, following standard gradient-based meta-learning algorithms like MAML~\cite{maml}, we adopt the single step update. This helps to reduce the optimization time as well as the requirements for large memory.

\textbf{Training Efficiency}. The double loop optimization paradigm of meta-learning has its intrinsic limitation. However, MoNA mitigates this issue in practice and achieves comparable efficiency with ORCA, as shown in table~\ref{tab:time}. The reason is two fold: first, we adopt single step update as discussed above to reduce the inner-loop optimization time. Second, MoNA requires only 5 epochs of target embedder training to achieve satisfactory performance, which is significantly lesser than ORCA, which requires more than 60 epochs in the first stage. Therefore, although MoNA cost longer time in training one epoch, the overall training time is comparable between the two methods.  

\section{Related Work}
\label{sec:related}
\vspace{-5pt}
\subsection{Cross-domain Transfer}
\vspace{-5pt}
Transfer learning is extensively investigated in the topic of domain shift~\cite{survey}. 
Domain Adaptation (DA)~\cite{ben2010theory,DAN-PAMI,DANN,GDCAN} studies the knowledge transfer between source and target domain that differ in data distribution yet share the same task. finetuning~\cite{transferable,L2SP,co-tuning,theory_ftvslp,borlan} is another powerful pipeline that enables knowledge transfer from large source dataset to different downstream tasks. It initializes the target model for downstream tasks using weights pretrained on massive source data, achieving higher performances than training the model from scratch. Our work belongs to the second transfer pipeline, yet addressing the more challenging situation where knowledge required for solving target tasks may not be quite aligned with source.

\vspace{-5pt}
\subsection{Cross-modality Transfer}
\vspace{-5pt}
Cross-modal transfer extends the boundary of transfer learning from same modality to different modalities~\cite{orca}. The possibility of leveraging model pretrained on one modality to benefit tasks on other modalities gains increasing attention in recent years~\cite{lift,fpt,reid2022can,llm_pattern,lm4ve}. One stream of works focuses on the knowledge reuse and transfer from source pretrained models. FPT~\cite{fpt} empirically shows that a pretrained language model (PLM) can benefit a variety of non-language downstream tasks, whereas other works transfer PLM to specific modalities~\cite{lm4tabular,lm4ve,reid2022can}. ORCA proposes a general workflow for cross-modal transfer that first aligns the data distribution between source and target embeddings and then finetunes the source model to adapt to target modality. Another line of researches aims to learn a general model for several modalities. Meta-transformer~\cite{metaTrans} proposes to design modality-specific embedders while keeping a multimodal pretrained model like CLIP~\cite{clip} frozen as the unified backbone. OneLLM~\cite{onellm} further adds projection layers on top of the pretrained encoder, interfacing it to large language models.

Despite the empirical success of previous works, we still lack understanding of the reason behind these success. As the target modalities can be greatly diverse, the common assumption that knowledge from source modality is beneficial to all target modalities requires further examination. In this work, we formalize the knowledge discrepancy between modalities as an interpretation to the actual feasibility of knowledge transfer between certain source to target modality, and propose a novel method to reduce modality knowledge misalignment.

\vspace{-5pt}
\subsection{Meta-Learning}
\vspace{-5pt}
Our work leverages the optimization-based meta-learning approaches that use bi-level optimization to embed learning procedures like gradient descent into the meta-optimization problem~\cite{survey_meta,imaml}. MAML~\cite{maml} is a representative work alone this line of meta-learning. It learns parameters that can serve as a general initialization to solve new downstream tasks with high efficiency. The method mimics the learning process on new tasks in the inner-loop, and updates the outer-loop parameters according to the final performances of each new task within the inner-loop. The spirit of such idea also inspires researches in domain generalization~\cite{metaDG,metareg,metaDG2,feature_critic,metaDG_med}, where the inner-loop simulates the model's generalization in unseen target domains.

\vspace{-5pt}
\section{Conclusion and Discussion}
\vspace{-5pt}
In this work, we empirically reveal the connection between the modality knowledge discrepancy and the effectiveness of cross-modal transfer. We provide interpretation of such discrepancy in terms of the divergence between conditional distributions. We further propose MoNA, a meta-learning based method to align source and target modality knowledge and improve from existing cross-modal transfer methods. Extensive experiments on two benchmarks with various modalities validate our approach.

Based on our formulation of modality knowledge discrepancy, future work may involve evaluating different source modalities and pretrained models to find the most transferable source model to the given target task.

\nocite{huggingface,pytorch,covariate-shift,bert,rereading}

\vspace{-5pt}
\section*{Acknowledgement}
\vspace{-5pt}
This paper was supported by the National Natural Science Foundation of China (No. 62376026), Beijing Nova Program (No. 20230484296) and Kuaishou.

\section*{Impact Statement}
The investigation of cross-modal transfer learning in this paper reveals the connection between modality knowledge discrepancy and cross-modality transfer effectiveness, which provides new insights to the applications when finetuning vision or language pretrained models to a variety of tasks such as PDE solving, cardiac disease prediction and hand gesture recognition. Additionally, the proposed meta-learning-based modality knowledge alignment method has the potential to enhance cross-modal transfer performance in diverse fields mentioned above and improve the real-world utility of deep neural networks. 

\bibliography{reference}
\bibliographystyle{icml2024}

\newpage
\appendix
\onecolumn
\section{Appendix}
\subsection{Benchmark Introduction and Training Details}
We summarize the details of ten tasks in NAS-Bench-360 in table~\ref{tab:sup_nasbench}. The ten tasks can be divided into three groups: 2D point prediction (classification), 2D dense prediction and 1D classification.

CIFAR-100~\cite{cifar100} in a standard classification task on natural images. Spherical~\cite{spherical} classifies spherical projections of the CIFAR-100 images which simulate distorted image signals. NinaPro~\cite{ninapro} moves away from the image modality to classify hand gestures indicated by electromyography signals. FSD50K~\cite{fsd} is an audio classification task originated from the larger Freesound dataset~\cite{freesound} with spectrogram as input and multiple labels as output. 

Darcy-Flow~\cite{fno} is a regression task for learning a map from the initial conditions of Partial Differential Equations (PDEs) to the solution at a later time-step. PSICOV~\cite{psicov} predicts the inter-residual distance of a small set of protein structures. Cosmic~\cite{cosmic} is the last dense prediction task in benchmark, aiming to identify comsic ray contamination in the images collected from the Hubble Space Telescope.

ECG~\cite{ecg} is the classification task on electrocardiogram signals that is frequently used in heart disease diagnosis. Satellite~\cite{satellite} is the classification of land cover type giving the satellite image time series as inputs. DeepSEA~\cite{deepsea} predicts the functional effects from genetic sequences and makes prediction among 36 categories of chromatin protein behavior. Please find more detailed description of these tasks in the original paper~\cite{nas-bench-360}.

The training configurations for vanilla finetuning (MoNA's second stage) for each task basically follow the setups in ORCA~\cite{orca} and is summarized in table~\ref{tab:sup_nasconfig}. For the first stage, we uniformly adopt AdamW~\cite{adamw} with learning rate 3e-5 and weight decay 0.1 since we find it works reasonably well on all tasks. We warmup the embedder with ten epochs before moving to the second stage.
\begin{table}[htbp]
  \centering
  \caption{Introduction of the ten tasks in NAS-Bench-360.}
  \resizebox{0.98\textwidth}{!}{%
    \begin{tabular}{lccccccccccccccccccccc}
    \toprule
    {} & {CIFAR100} & {Spherical} & {NinaPro} & {FSD50K} & {DarcyFlow} & {PSICOV} & {Cosmic} & {ECG} & {Satellite} & {DeepSEA} \\
    \midrule
    {\# training data} & {60K} & {60K} & {3956} & {51K} & {1.1K} & {3606} & {5250} & {330K} & {1M} & {250K} \\
    {Input shape} & {2D} & {2D} & {2D} & {2D} & {2D} & {2D} & {2D} & {1D} & {1D} & {1D} \\
    \midrule
    {Output type} & {Point} & {Point} & {Point} & {Point} & {Dense} & {Dense} & {Dense} & {Point} & {Point} & {Point} \\
    {\# classes} & {100} & {100} & {18} & {200} & {--} & {--} & {--} & {4} & {24} & {36} \\
    {Loss} & {CE} & {CE} & {LpLoss} & {MSELoss} & {BCE} & {FocalLoss} & {BCE} & {CE} & {CE} & {BCE} \\
    \midrule
    {\multirow{2}[2]{*}{Expert network}} & {\multirow{2}[2]{*}{DenseNet-BC}} & {\multirow{2}[2]{*}{S2CN}} & {\multirow{2}[2]{*}{Attention Model}} & {\multirow{2}[2]{*}{VGG}} & {\multirow{2}[2]{*}{FNO}} & {\multirow{2}[2]{*}{DEEPCON}} & {\multirow{2}[2]{*}{deepCR-mask}} & {\multirow{2}[2]{*}{ResNet-1D}} & {\multirow{2}[2]{*}{ROCKET}} & {\multirow{2}[2]{*}{DeepSEA}} \\
    {} & {} & {} & {} & {} & {} & {} & {} & {} & {} & {} \\
    \bottomrule
    \end{tabular}%
  }
  \label{tab:sup_nasbench}%
\end{table}%

\begin{table}[htbp]
  \centering
  \caption{Configuration of hyper-parameters and optimizers used in NAS-Bench-360.}
  \resizebox{0.98\textwidth}{!}{%
    \begin{tabular}{lccccccccccccccccccccc}
    \toprule
    {} & {CIFAR100} & {Spherical} & {NinaPro} & {FSD50K} & {Darcy Flow} & {PSICOV} & {Cosmic} & {ECG} & {Satellite} & {DeepSEA} \\
    \midrule
    {Batch Size} & {32} & {32} & {32} & {32} & {4} & {1} & {4} & {4} & {16} & {16} \\
    \midrule
    {Epoch} & {60} & {60} & {60} & {100} & {100} & {10} & {60} & {15} & {60} & {13} \\
    \midrule
    {Grad. Accum.} & {32} & {4} & {1} & {1} & {1} & {32} & {1} & {16} & {4} & {1} \\
    \midrule
    {Optimizer} & {SGD} & {AdamW} & {Adam} & {Adam} & {AdamW} & {Adam} & {AdamW} & {SGD} & {AdamW} & {Adam} \\
    \midrule
    {Learning Rate} & {1.00E-04} & {1.00E-04} & {1.00E-04} & {1.00E-04} & {1.00E-03} & {5.00E-06} & {1.00E-03} & {1.00E-06} & {3.00E-05} & {1.00E-05} \\
    \midrule
    {Weight Decay} & {1.00E-03} & {1.00E-01} & {1.00E-05} & {5.00E-05} & {5.00E-03} & {1.00E-05} & {0.00E+00} & {1.00E-01} & {3.00E-06} & {0.00E+00} \\
    \bottomrule
    \end{tabular}%
  }
  \label{tab:sup_nasconfig}%
\end{table}%

On PDEBench, we evaluate all tasks except 2D and 3D Navier-Stokes Equations which are too computational expensive. We provide simple introduction to each PDE and refer more details to the original paper~\cite{pdebench}.

1D Advection equation models pure advection behavior without non-linearity, with parameter $\beta$ informing the constant advection speed. 1D Burgers' equation models the non-linear behavior and diffusion process in fluid dynamics. The parameter $\nu$ is the diffusion coefficient which is assumed constant. 1D Diffusion-Reaction equation combines a diffusion process and a rapid evolution from a source term, where two parameters $\nu,\rho$ control the degree of combination. 1D Diffusion-Sorption equation models a diffusion process which is retarded by a sorption process. The equation is applicable to real world scenarios. 1D compressible Navier-Stokes equation describes the dynamics of compressible fluid, where $\eta$ and $\zeta$ are the shear and bulk viscosity, respectively.

2D Darcy-Flow equation describes a steady-state solution of the flow dynamics over the unit square. The force term $f(x)$ is simplified as the constant $\beta$ and it changes the scale of the solution. 2D shallow-water equations are derived from the general Navier-Stokes equation that presents a suitable framework for modelling free-surface flow problems. 2D Diffusion-Reaction equation extends the 1D equation by considering two non-linearly coupled variables.This task serves as a challenging problem since the coupling is non-linear and its real world application is huge.

\begin{table}[htbp]
  \centering
  \caption{Introduction of eight tasks evaluated on PDEBench.}
  \resizebox{0.96\textwidth}{!}{%
    \begin{tabular}{lcccccccc}
    \toprule
    {} & {Advection} & {Burgers} & {Diffusion-Reaction} & {Diffusion-Sorption} & {Navier-Stokes} & {Darcy-Flow} & {Shallow-Water} & {Diffusion-Reaction} \\
    \midrule
    {Input shape} & {1D} & {1D} & {1D} & {1D} & {1D} & {2D} & {2D} & {2D} \\
    \midrule
    {Output type} & \multicolumn{8}{c}{Dense} \\
    \midrule
    {Resolution} & {1024} & {1024} & {1024} & {1024} & {1024} & {128 * 128} & {128 * 128} & {128 * 128} \\
    \midrule
    {Parameters} & {$\beta=0.4$} & {$\nu=1.0$} &  {$\nu=0.5,\rho=1.0$} & {--} & {$\eta=1.0,\zeta =1.0$} & {$\beta=0.1$} & {--} & {--} \\
    \midrule
    {Loss} & \multicolumn{8}{c}{Normalized Root Mean Squared Errors (nRMSEs)} \\
    \bottomrule
    \end{tabular}%
  }
  \label{tab:addlabel}%
\end{table}%

\begin{table}[htbp]
  \centering
  \caption{Configuration of hyper-parameters and optimizers used in NAS-Bench-360}
  \resizebox{0.96\textwidth}{!}{%
    \begin{tabular}{lccccccccccccccccc}
    \toprule
    {} & {Advection} & {Burgers} & {Diffusion-Reaction} & {Diffusion-Sorption} & {Navier-Stokes} & {Darcy-Flow} & {Shallow-Water} & {Diffusion-Reaction} \\
    \midrule
    {Batch Size} & {4} & {4} & {4} & {4} & {4} & {4} & {4} & {4} \\
    \midrule
    {Epoch} & {200} & {200} & {200} & {200} & {200} & {100} & {200} & {200} \\
    \midrule
    {Grad. Accum.} & {1} & {1} & {1} & {1} & {1} & {1} & {1} & {1} \\
    \midrule
    {Optimizer} & {Adam} & {Adam} & {SGD} & {AdamW} & {AdamW} & {AdamW} & {AdamW} & {Adam} \\
    \midrule
    {Learning Rate} & {1.00E-04} & {1.00E-05} & {1.00E-03} & {1.00E-04} & {1.00E-04} & {1.00E-04} & {1.00E-04} & {1.00E-04} \\
    \midrule
    {Weight Decay} & {1.00E-05} & {1.00E-05} & {1.00E-05} & {0} & {1.00E-03} & {1.00E-05} & {0} & {1.00E-03} \\
    \bottomrule
    \end{tabular}%
  }
  \label{tab:addlabel}%
\end{table}%

\subsection{Detailed Explanation of the Model Architecture}
In this section we provide a detailed explanation of the modality-specific embedders and predictors. Note that our implementation of these modules follows exactly the designs in ORCA.

The structure of the modality-specific embedder depends on whether the task is 2D or 1D.

\begin{itemize}
  \item For 2D tasks, the embedder consists of a linear projection layer and a LayerNorm operation. For any input data with size $C\times H\times W$, where $C$,$H$ and $W$ are channels, height and width, we first resize it to $C\times 224^2$ and divide it into $N$ patches of size $C \times 4^2$. Then the linear projection layer maps each patch into a token of size $128$ and the 
  LayerNorm operation is applied on all the projected patches. Therefore, the embedder can be formulated as a function $e_{2D}: \mathbb R^{N\times 16C}\to\mathbb R^{N\times 128}$.
  \item For 1D tasks, the embedder consists of a linear projection layer, a LayerNorm operation and learnable positional embeddings. For any input data with size $C\times L$ where $C$ and $L$ are channels and sequence length respectively, we first divide it into $N$ patches of size $C \times \frac{L}{N}$. Then the linear projection layer maps each patch into a token of size $768$ and the LayerNorm operation is applied on all the projector patches. Finally, the positional embeddings are added to the patches. Therefore, the embedder can be formulated as a function $e_{1D}: \mathbb R^{CL}\to\mathbb R^{768N}$.
\end{itemize}

The structure of the modality-specific predictor depends on whether the task is classification or dense prediction.

\begin{itemize}
  \item For classification tasks, the predictor consists of an average pooling layer and a linear projection layer. The average pooling layer averages the dense feature map of size $N'\times d$ to produce feature of size $d$, then the linear projection layer maps the feature to logits of size $K$, where $d$ and $K$ represent feature dimension and the number of categories. Therefore, the predictor can be formulated as a function $h_{c}: \mathbb R^{N'd} \to \mathbb R^K$.
  \item For dense prediction tasks, the predictor consists of a linear projection layer, a pixel rearrangement operation and two adaptive pooling layers. The linear projection layer takes the dense feature map of size $7^2\times d$ as input and output a new feature of size $7^2 \times 3072$, which is then reorganized into shape $224^2 \times 3$. Next, two pooling operations are applied sequentially, turning the feature size from $3 \times 224^2$ to $K \times 224^2$ and finally $K \times H \times W$ which is in accordance with the input spatial dimension. Therefore, the predictor can be formulated as a function $h_{d}:\mathbb R^{49d}\to\mathbb R^{KHW}$.
\end{itemize}

\subsection{Theoretical Foundation for the Meta-Learning Objective}
We provide a preliminary theoretical analysis of our meta-learning objective. As in our paper, we denote the target embedder parameter as $\phi_e$ and the pretrained encoder parameter as $\theta_f$. The objective of MoNA is:
\begin{equation}
  \min_{\phi_e} \mathcal L_{inner}(\theta_f, \phi_e )+\lambda \mathcal L_{outer}(\theta_f-\alpha\mathcal \nabla_{\theta_f} L_{inner}(\theta_f,\phi_e)).
\end{equation}
We then get the approximation of the objective using Taylor expansion as:
\begin{equation}
  \min_{\phi_e} \mathcal L_{inner}(\theta_f,\phi_e)+\lambda\mathcal L_{outer}(\theta_f)+\lambda\nabla\mathcal L_{outer}(\theta_f)\cdot(-\alpha\nabla\mathcal L_{inner}(\theta_f,\phi_e)).
\end{equation}
Since it is the optimization problem of $\phi_e$, the second term is neglected and we obtain the final form as:
\begin{equation}
  \min_{\phi_e}\mathcal L_{inner}(\theta_f,\phi_e)-\lambda\alpha(\nabla\mathcal L_{outer}(\theta_f))\cdot(\nabla\mathcal L_{inner}(\theta_f,\phi_e)).
\end{equation}
We see that the first term directly minimizes the target task loss, whereas the second term maximizes the dot product of the source loss gradient and the target loss gradient, both with respect to $\theta_f$. In other words, the second term updates the target embedder in a way that makes the gradient direction of the target task loss align with the gradient direction of the source modality loss. In the optimal situation where the two gradients point to the same direction, finetuning the encoder $\theta_f$ using target loss will maximally maintain the source knowledge. With this theoretical support, we argue the optimality of the target embedder update objective in terms of achieving cross-modal knowledge alignment.

\subsection{Approximation Algorithm for Computing Modality Knowledge Discrepancy}
We explain the algorithm used to approximate the modality knowledge discrepancy in definition 2.2. The results is shown in Figure 4. For each target sample $(\x^t_i, y^t_i)$, we use the one-hot embedding of its label as the target conditional distribution $p(y^t|\x^t_i)$. For the source conditional distribution, we leverage the complete source pretrained model (with the original classifier trained on ImageNet) to compute the logits $\boldsymbol{z}^s_i$. We then simplify the searching for optimal subset $\mathcal{B}$ in the definition as a random category selection. Number of the selected source classes are equal to the number of the target categories being compared. With the subset selected, we consider the maximum logit value within the subset and assign the source category as $y^s_{i,\mathcal{B}} = \arg\max_{k\in\mathcal{B}} [\boldsymbol{z}^s_i]_k$, and we also use the one-hot embedding of source predicted category to model the source conditional distribution $p(y^s_{\mathcal{B}}|\x^t_i)$. To this end, the discrepancy between two conditional distribution can be simplified as 
\begin{equation}
d(p(y^s_{\mathcal{B}}|\x^t_i),p(y^t|\x^t_i)) = \mathbb{I}_{[y^s_{i,\mathcal{B}}\neq y^t_i]},
\end{equation}
Therefore, we can compute the modality knowledge discrepancy as
\begin{equation}\label{Eq:d}
  D(\mathcal{M}^s,\mathcal{M}^t) = \frac{1}{n_t} \sum_{i=1}^{n_t} d(p(y^s_{\mathcal{B}}|\x^t_i),p(y^t|\x^t_i)).
\end{equation}

Still, we need to find the optimal permutation $\pi$ that matches source and target categories one-to-one. Since it is too computational expensive to iteration through all the permutations, we opt to randomly permute the target label indexes. Therefore in practice, we conduct random experiment for 100,000 times. Each time we randomly select subset of the source and randomly permute the target label indexes. We compute the modality knowledge discrepancy using Eq.~\eqref{Eq:d}, and the final discrepancy is the minimum value during the whole process. Alg.~\ref{alg:apprx} summarizes the complete process.

\begin{algorithm}[t!]
  \caption{Approximation Algorithm for Modality Knowledge Discrepancy}\label{alg:apprx}
  \begin{algorithmic}[1]
  \STATE {\bfseries Input:}
      Source pretrained model $g_{\tha^\Sm_0}$; Target data $\{\x^t_i,y^t_i\}$; Target task category number $K$; Maximum experiments $I$.
  \STATE {\bfseries Output:}
      Modality Knowledge Discrepancy $D(\mathcal{M}^s,\mathcal{M}^t)$.

  \STATE Compute the source logit for each target sample: $\boldsymbol{z}^s_i = g_{\tha^\Sm_0}
  (\x^t_i)$.
  \STATE $min_D \leftarrow 1$.
  \FOR{$exp=1,2,\cdots,I$}
  \STATE Randomly select a subset $\mathcal{B}$ of class index from source class indexes with the size equals to target class number, $\vert\mathcal{B}\vert = K$.
  \STATE $y^s_{i,\mathcal{B}} \leftarrow \arg\max_{k\in\mathcal{B}} [\boldsymbol{z}^s_i]_k$
  \STATE Randomly shuffle the index of target labels: $y^t \leftarrow \pi(y^t)$, (e.g., $\pi(1)=2,\pi(2)=1$.)
  \STATE Compute $D_{exp}(\mathcal{M}^s,\mathcal{M}^t)$ using Eq.~\eqref{Eq:d}.
  \STATE $min_D \leftarrow \min(D_{exp}, min_D)$.
  \ENDFOR
  \STATE Return $min_D$.
  \end{algorithmic}
\end{algorithm}

\subsection{Cross-Modality Transfer Against Cross-Modality Knowledge Distillation}
Cross-modal Knowledge Distillation~\cite{mfh} is an alternative paradigm for cross-modality knowledge sharing, and is proven to be useful on diverse applications including video representation learning~\cite{xkd}, action recognition~\cite{garcia2018modality,dai2021learning}, lip reading~\cite{ren2021learning,asr}, depth~\cite{gupta2016cross}, sound~\cite{soundnet} and etc. Specifically, XKD~\cite{xkd} explores leveraging Maximum Mean Discrepancy to align video and image modality, ASR~\cite{asr} proposes a novel Connectionist Temporal Classification loss that enables learning sequence-to-sequence tasks without the need for explicit alignment of training targets to input frames, and Augmented RGB~\cite{dai2021learning} also investigates the sequence-to-sequence knowledge distillation framework using a contrastive strategy.

A common thread among these methodologies is their reliance on paired data or multimodal representations of identical data points. This prerequisite, however, may not always be feasible or accessible in certain modalities, such as those involving Partial Differential Equations (PDEs) or protein structure prediction, thereby limiting their applicability.

In contrast, cross-modal transfer learning emerges as a more versatile and inclusive framework for knowledge sharing across modalities, primarily because it eschews the need for paired data, thereby casting a wider net in terms of application potential. However, this flexibility comes at the cost of an increased risk of ineffective transfer, particularly when faced with substantial modality knowledge discrepancies and in the absence of paired data to serve as a bridge between the disparate modalities.

Therefore, one of the primal purposes of this work is to describe the extent of modality discrepancy systematically. We hope that our effort can motivate research in modality discrepancy, which would eventually provide guidelines for better cross-modal knowledge transfer.

\subsection{Limitation}
Here we discuss a few limitations of our work as well as potential solution towards these limitations.

Firstly, our experimental framework adheres to the protocols established by ORCA, which involves utilizing CIFAR10 and CoNLL-2003 as surrogate source datasets for the vision and language modalities, respectively. This approach, while facilitating a direct comparison with established benchmarks, introduces a limitation in that the choice of surrogate datasets might influence the outcomes of cross-modal transfer learning. The potential variability in transfer performance attributed to different source datasets is a factor that our current analysis does not account for, presenting a critical area for future exploration.

Secondly, in the implementation of MoNA, we opted for a simplistic yet effective strategy akin to Model-Agnostic Meta-Learning (MAML), using a single-step update within the inner loop to balance the performance and computational cost. Noticed that with recent advancements in gradient-based meta-learning proposing improved algorithms for inner-loop optimization, we are committed to actively exploring these methods to improve the algorithm of MoNA.


\end{document}